\documentclass[letterpaper]{article} 

\usepackage{aaai24}  

\usepackage[utf8]{inputenc} 
\usepackage[T1]{fontenc}    
\usepackage{booktabs}  
\usepackage{amsfonts}       
\usepackage{nicefrac}       
\usepackage{microtype}      
\usepackage{xcolor}         
\usepackage{amsmath}
\usepackage{verbatim}
\usepackage[symbol]{footmisc}
\usepackage{subcaption}
\usepackage{makecell}
\usepackage{dsfont}

\usepackage{times}  
\usepackage{helvet}  
\usepackage{courier}  
\usepackage[hyphens]{url}  
\usepackage{graphicx} 
\usepackage{natbib}  
\usepackage{caption} 
\title{A Baseline Analysis of Reward Models' Ability To Accurately Analyze Foundation Models Under Distribution Shift}

\author{Will LeVine\footnote{These authors contributed equally.}\footnote{Corresponding email: levinewill@icloud.com}, Benjamin Pikus$^*$,  Anthony Chen, Sean Hendryx}
\affiliations{Scale AI}

\begin{document}

\maketitle

\begin{abstract}

Foundation models, specifically Large Language Models (LLMs), have lately gained wide-spread attention and adoption. Reinforcement Learning with Human Feedback (RLHF) involves training a reward model to capture desired behaviors, which is then used to align LLMs. These reward models are additionally used at inference-time to estimate LLM response adherence to those desired behaviors. However, there is little work measuring how robust these reward models are to distribution shifts. In this work, we evaluate how reward model performance - measured via accuracy and calibration (i.e. alignment between accuracy and confidence) - is affected by distribution shift. We show novel calibration patterns and accuracy drops due to OOD prompts and responses, and that the reward model is more sensitive to shifts in responses than prompts. Additionally, we adapt an OOD detection technique commonly used in classification to the reward model setting to detect these distribution shifts in prompts and responses.
  
\end{abstract}

\section{Introduction}
Large Language Models, such as ChatGPT, have lately greatly increased in fame and usage. These models are typically finetuned via RLHF using reward models in order to align model responses towards rewarded behaviors \citep{christiano2017deep, ziegler2019fine}. Beyond being used for training, these reward models can be used to assess how well LLM responses adhere to those rewarded behaviors. However, inference distributions are not always stationary and sometimes distribution shifts occur where test-time examples are far from the training set or in low-density pockets of the training set (e.g. users ask questions that are unlike most of the data on which the foundational model was trained). Under distribution shifts, classification models are widely known to degrade in terms of performance and calibration \citep{NEURIPS2019_8558cb40}. This therefore opens the question of how reward models behave under distribution shift. In this paper, we therefore 
\begin{enumerate}
    \item Study the behavior of reward model ability to accurately assess LLMs under distribution shift. We show that reward model accuracy strictly degrades under distribution shifts in prompts and responses, with higher magnitude drops due to OOD responses.
    \item Study the behavior of reward model calibration. We show that: reward model calibration due to OOD prompts is relatively unaffected by distribution shift; while reward model calibration due to OOD responses follows a novel paradigm with excellent calibration far-OOD (even better than ID calibration) but poor calibration near-OOD due to overconfidence.
    \item Introduce a technique inspired by classification to detect responses and prompts that are far from the training set. This allows the identification of when reward models are unable to reliably analyze responses to prompts in terms of adherence to a rewarded behavior.
\end{enumerate}

\section{Related Works}
\subsection{Analyzing Reward Models Under Distribution Shift}
\citet{stepbystep} showed that the performance of reward models decreases under distribution shift across different STEM questions. And \citet{genies_reward} investigated reward model shifts concurrently to this work and found similar degradation in performance as a result of different types of shifts - although they studied such when prompts and responses were both shifted concurrently.

\subsection{OOD Detection In Reward Models}
\citet{liu2023good} presented methods to detect OOD prompts in LLMs (but not reward models). However, our aim is to detect both OOD responses and OOD prompts (not just OOD prompts), especially since we will show that distribution shifts in responses cause performance degradations in reward models moreso than distribution shifts in prompts; and their method relies on having access to internal model probabilities, which is not possible with closed source LLM’s.

\section{Preliminaries}
\subsection{Classification}
\label{problem_setup}
\label{section:classification_setup}
Let $X$ and $Y$ be the input and response random variables with realizations $x \in \mathbb{R}^D$ and $y \in \{1,2,...,C-1,C\}$, respectively, where $C$ is the number of output classes. Given learned logit function $\hat{L}^{clf} : \mathbb{R}^D \to \mathbb{R}^C$, the model prediction (including softmax) is $$\hat{f}^{clf}_c(x_i) = e^{\hat{L}^{clf}_c(x_i)} / \sum_{j=1}^Ce^{\hat{L}^{clf}_j(x_i)}$$ with confidence $$\hat{p}(x_i, \hat{f}^{clf}) = \underset{c}{\mathrm{max}}\hat{f}^{clf}_c(x_i)$$

\subsubsection{Evaluating Classification Models}

\label{section:classification:accuracy}
Given unseen $D^{test} = \{(x_i, y_i)\}_{i=1}^N$, we evaluate the classification performance of $\hat{f}^{clf}$ using classification accuracy: $$\mathrm{acc^{clf}}(D^{test}, \hat{f}^{clf}) = \dfrac{1}{N}\sum_{i=1}^{N}\mathds{1}[\underset{c}{\mathrm{argmax}}\hat{f}^{clf}_c(x_i) = y_i]$$

We evaluate the calibration of classification models as the alignment of confidence and accuracy. As an example from \citet{guo2017calibration}, given a set of 100 predictions with confidences of $0.8$, we would hope that 80 of these predictions would be correctly classified. If so, we would consider the model to be \textit{calibrated}. Let $D^{test}_p = \{(x_i, y_i) \in D^{test} \hspace{1mm} \text{s.t.} \hspace{1mm} \hat{p}(x_i, \hat{f}^{clf}) = p\}$. Formally, a model is calibrated if $$\text{acc}^{\text{clf}}(D^{test}_p, \hat{f}^{clf}) = p\hspace{1mm}\forall\hspace{1mm}p \in [0, 1]$$ We further note as in \citet{guo2017calibration} that the accuracy in this  equation cannot be computed on a single sample, since an accuracy is computed on a set of examples rather than a single sample. Hence the need for Expected Calibration Error ($\textit{ECE}^{clf}$), which empirically approximates alignment of confidence and accuracy. To calculate $\textit{ECE}^{clf}$, points are grouped by their predicted confidence scores into $M$ equally spaced bins. These are formalized as $B_m$ denoting a bin containing the test samples that falls into the interval $I_m = \big((m - 1)/M, m/M\big]$, for $m = 2, \dots, M$, and $I_1 = [0, \frac{1}{M}]$. The true accuracy in $B_m$ is $\text{acc}(B_m, \hat{f}^{clf})$ and the \textit{estimated} accuracy (i.e. average confidence) in $B_m$ is $1/|B_m|\sum_{(x_i, y_i) \in B_m}\hat{p}(x_i, \hat{f}^{clf})$, which we write in short-hand as $\hat{p}(B_m, \hat{f}^{clf})$. For all experiments, we let $M = 10$, as is standard \citep{levine2023enabling, guo2017calibration, kull2019beyond, rajendran2019accurate}. $\textit{ECE}^{clf}$ is then calculated as $$\text{ECE}^{clf} = \sum_{m = 1}^M \dfrac{|B_m|}{|D|}\bigg\lvert\hat{p}(B_m, \hat{f}^{clf}) - \text{acc}(B_m, \hat{f}^{clf}) \bigg\rvert$$

\subsubsection{The Effects of Distribution Shift On Classification Performance and Calibration}
Classification accuracy is widely known to deteriorate under distribution shift. Additionally, ECE increases, signaling predictions becoming miscalibrated \citep{NEURIPS2019_8558cb40}. Therefore, towards the safety and reliability of classification models, Out-of-Distribution Detection  - or ``OOD Detection" - aims to identify inference samples far from the training set.

\subsubsection{Out-of-Distribution Detection in Classification}
$D^{test}_{out}$ in classification is typically defined as any dataset that is significantly different from $D^{train}$ \citep{huang2021importance, hsu2020generalized, liu2020energy, djurisic2022extremely, sun2021react, hendrycks2016baseline, liang2017enhancing, sun2022out, katz2022training, levine2024outofdistribution} - e.g. day vs. night. Out-of-Distribution Detection estimators in classification aim to define a score $S$ such that $S(x_{out})$ and $S(x_{in})$ are far from each other $\forall \hspace{1mm} x_{out} \in D_{out}^{test}, x_{in} \in D^{test}$. 

\paragraph{Detecting OOD Samples In Classification Via Energy Score}

A simple Out-of-Distribution score of a model with logit function $\hat{L}^{clf}$ on inference example $x_i$ is the Energy Score from \citet{liu2020energy}: $$S^{clf}(x_i, \hat{L}^{clf}) = -\log\sum_{c=1}^Ce^{\hat{L}^{clf}_c(x_i)}$$ These logits are trained such that $\hat{L}^{clf}_c$ increases as example $x_i$ more closely resembles the training examples of class $c$. Intuitively, this Energy Score therefore measures the similarity of inference example $x_i$ to the training examples of the $C$ training classes, and therefore the similarity to the training set. 

\subsection{Reward Models}
We now describe reward models, which we seek to analyze under distribution shift. 
\subsubsection{Reward Models Problem Setup}
Reward models measure the alignment of an LLM-generated response $r_i$ to a prompt $p_i$ in terms of adherence to a rewarded behavior. They do so by outputting a logit $\hat{L}^{rwd}(r_i, p_i)$ which is trained to be higher when the response to the prompt adheres more to the rewarded behavior (e.g. is more helpful or less harmful). During training, the reward model is trained on a prompt and two responses, where one of the responses is preferred to the other. 
We formalize this train set as $D^{train} = \{((p_i, (r^0_i, r^1_i)), l_i)\}_{i=1}^{N}$ , where $p_i$ is the prompt, $r_i^0$ and $r_i^1$ are the two responses, and $l_i \in \{0, 1\}$ denotes which of the two responses adheres more to the rewarded behavior (i.e. which is the preferred response). We define the confidence of $\hat{f}^{rwd}$ in its prediction identically to the calculation of confidence in classification models in Section \ref{section:classification_setup}, but treating the two responses as the classes. More specifically, confidence is the max softmax of the logits of the two prompt-response pairs; or, more formally: $$\hat{p}(x_i, \hat{f}^{rwd}) = \underset{j\in \{0, 1\}}{\mathrm{max}} \hspace{1mm} e^{\hat{L}^{rwd}(p_i, r_i^j)} / \sum_{j=0}^1e^{\hat{L}^{rwd}_j(p_i, r_i^j)}$$ 

\begin{table*}[t]
\centering
\begin{tabular}{llll}
\hline
Prompt Distribution      & Response Distribution         & Accuracy $\uparrow$         & ECE $\downarrow$\\ \hline
ID     & ID          &      72.3\%    & 14.53\% \\ 
OOD    & ID          &        70.29 $\pm$ 0.08\%   & 10.8 $\pm$ 0.21\%\\
ID     &  OOD          &       65.69 $\pm$ 0.52\%    & 20.03 $\pm$ 0.56\% \\
OOD     & OOD          &      64.44 $\pm$ 0.73\% & 19.8 $\pm$ 0.54\%\\
\hline
\end{tabular}
\centering
\caption{Comparison of reward model performances under lingual distribution shifts in prompt and response. Averages and standard deviations are taken across OOD language. $\uparrow$ means higher is better and $\downarrow$ means lower is better.
}
\label{table:accuracy_degradation}
\end{table*}

\begin{figure*}[t]
    \hspace{2em}
    \centering
    \begin{subfigure}{0.2\textwidth}
        \centering
        \includegraphics[height=0.9in]{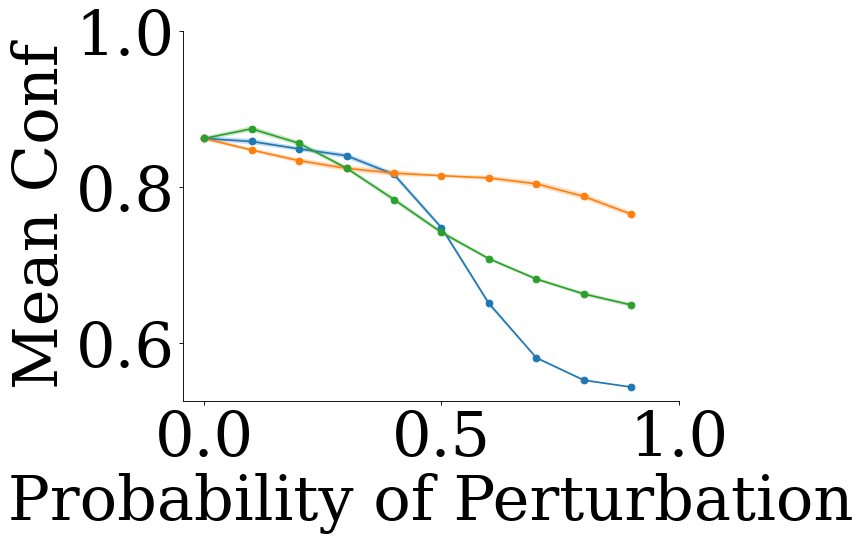}
    \end{subfigure}%
    ~ 
    \begin{subfigure}{0.33\textwidth}
        \centering
        \includegraphics[height=0.9in]{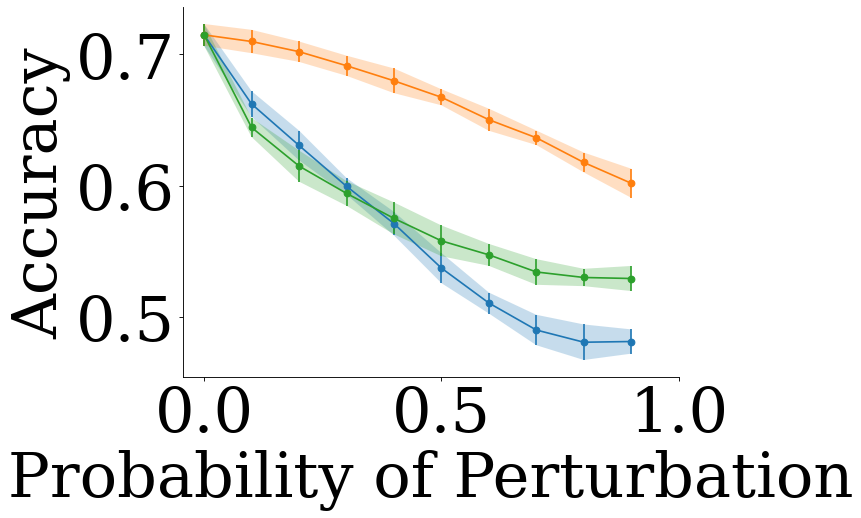}
    \end{subfigure}%
    ~ 
    \begin{subfigure}{0.33\textwidth}
        \centering
        \includegraphics[height=0.9in]{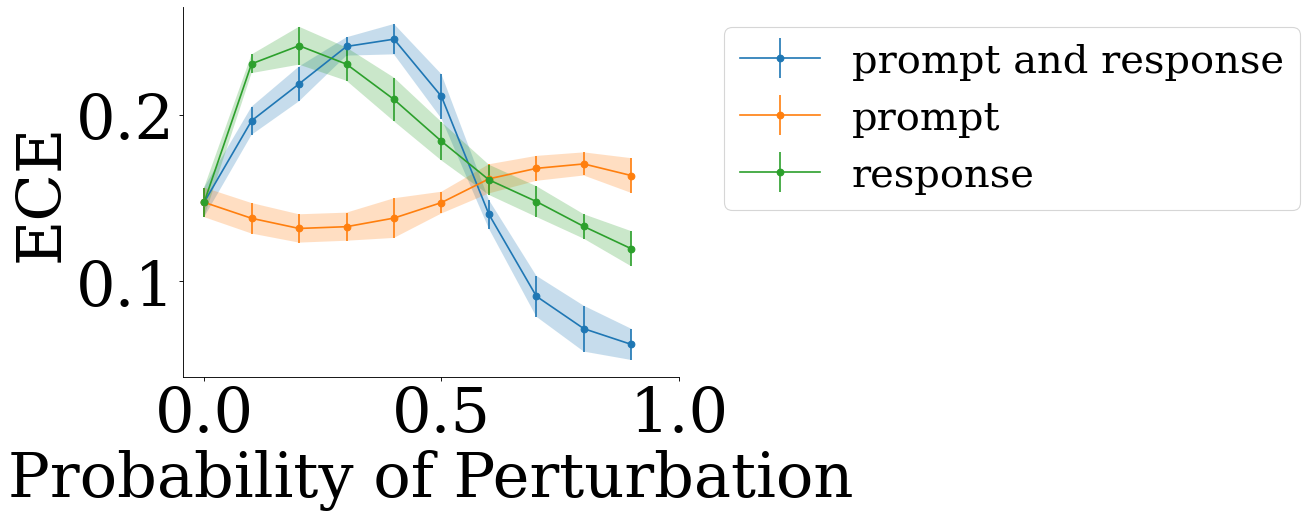}
    \end{subfigure}
    \caption{(\textbf{Left}) confidence of reward models, and performance of reward models under artificial distribution shift in terms of (\textbf{Middle}) accuracy - where higher is better - and (\textbf{Right}) \textit{ECE} - where lower is better. The legend indicates if the shift is in response, prompt, or both. Further right on the x-axis is further OOD. }
    \label{fig:dist_shift_acc_calibration}
\end{figure*}

\subsubsection{Evaluating Reward Models}
Ideally, the reward model is able to distinguish between the two responses to a prompt correctly in terms of which of the two responses adheres more to the rewarded behavior. We evaluate this preference selection performance via reward accuracy on unseen test dataset $D^{test}$ as follows: 
\begin{align*}
    &\mathrm{acc^{rwd}}(D^{test}, \hat{L}^{rwd}) \\= &\dfrac{1}{N}\sum_{i=1}^N\mathds{1}[\hat{L}^{rwd}(p_i, r^{l_i}_i) > \hat{L}^{rwd}(p_i, r^{1 - {l_i}}_i)
\end{align*}

To evaluate the calibration performance of reward models, we define $\textit{ECE}^{\textit{rwd}}$ identical to $\textit{ECE}^{\textit{clf}}$ but using $\mathrm{acc}^{\mathrm{rwd}}$ instead of $\mathrm{acc}^{\mathrm{clf}}$.

\section{Reward Model Performance Under Distribution Shift}

In all experiments, we use OpenAssistant's  \citep{debertareward} {\fontfamily{qcr}\selectfont 
 deberta-v3-large-v2} \citep{he2021deberta} as the reward model, evaluated on Summarize From Feedback \citep{stienon2020learning}, where each response attempts to well-summarize the prompt. 

\label{sec:performance_under_shift}

\subsection{Natural Distribution Shift}

In Table \ref{table:accuracy_degradation}, we present the accuracy and ECE of the reward model when the prompts and responses are ID and OOD. Specifically, ID prompts and responses are in English, similar to its training set Summarize From Feedback \citep{stienon2020learning}. In contrast, OOD prompts and responses are created by translating the English prompts and responses into French, Spanish, and German. OPUS-MT models \citep{opusmt} were used to perform all translation. We run all of our Table \ref{table:accuracy_degradation} experiments with OOD being entirely one language at a time, then present the average. Detailed results per-language can be found in Appendix Section \ref{sec:per_language_res}. We show that the accuracy of reward models significantly lower on OOD responses and prompts, with drops in accuracy significantly greater due to distributions shifts in responses than distribution shifts in prompts. We also show similar findings for calibration. But, interestingly, OOD prompts with ID responses result in \textit{improved} calibration over cases where both prompts and responses are ID - although this improvement is very slight, meaning the reward model is largely unaffected by OOD prompts in terms of calibration.  

We note that this form of distribution shift is coarse, and does not allow very granular analysis of the performance of reward models under distribution shift stratified by the magnitude of that shift. Therefore, to present a more granular analysis, we induce artificial distribution shifts of varying magnitudes and study the behavior of reward models under these shifts below in Section \ref{sec:artificial}.

\begin{table*}
\centering
\begin{tabular}{llll}
\hline
Prompt Distribution      & Response Distribution         & AUROC $\uparrow$         & FPR@95 $\downarrow$\\ \hline
ID     &  OOD          &       72.84 $\pm$ 1.32\%  &       71.46 $\pm$ 1.58\%  \\ 
OOD    & ID          &        63.11 $\pm$ 1.45\%  &       71.5 $\pm$ 1.92\%\\
OOD     & OOD          &      77.12 $\pm$ 2.11\% &      60.21 $\pm$ 3.31\% \\
\hline
\end{tabular}
\centering
\caption{OOD detection results under lingual distribution shift. $\uparrow$ means higher is better and $\downarrow$ means lower is better.
}

\label{table:natural_detection}
\end{table*}

\begin{figure*}[t]
    \centering
    \begin{subfigure}{0.5\textwidth}
        \centering
        \includegraphics[height=1.3in]{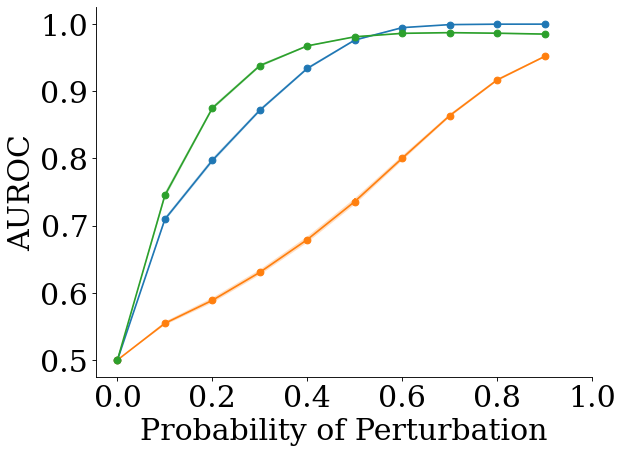}
    \end{subfigure}%
    ~ 
    \begin{subfigure}{0.5\textwidth}
        \centering
        \includegraphics[height=1.3in]{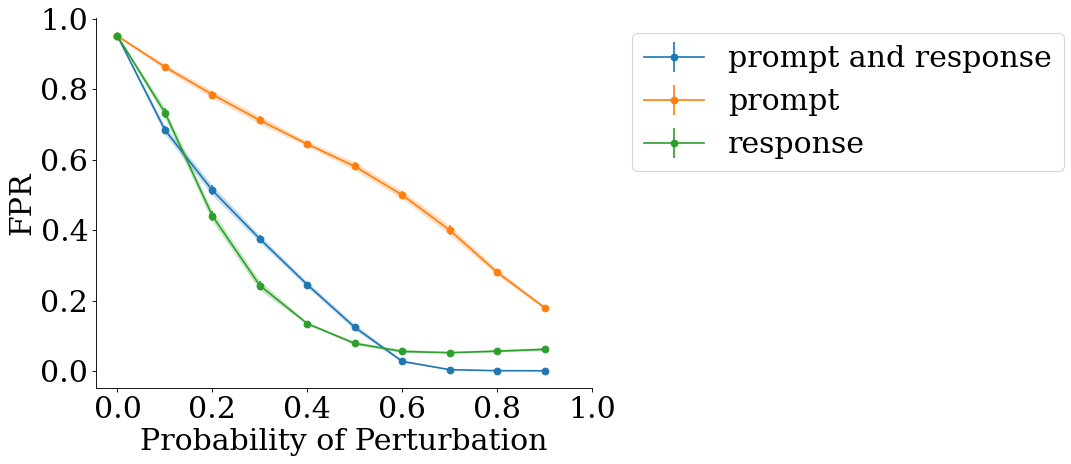}
    \end{subfigure}
    \caption{Performance of Energy Score in detecting artificial distribution shifts in prompts and responses, measured in terms of (\textbf{Left}) AUROC - where higher is better - and (\textbf{Right}) FPR@95 - where lower is better. The legend indicates if the shift is in response, prompt, or both. Further right on the x-axis is further OOD.}
    \label{fig:ood_detection}
\end{figure*}

\subsection{Artificial Distribution Shift}

\label{sec:artificial}

\subsubsection{Inducing An Artificial Distribution Shift}
We artificially induce distribution shift by perturbing words with some probability (where the perturbation is either an insertion, deletion, or replacement with a random word from the same language). A higher probability means a larger distribution shift. This induces distribution shift because these perturbations cause the prompts and responses to be more non-sensical - and therefore more dissimilar to the prompts and responses in the training set. Example perturbations can be found in Appendix Section 
\ref{sec:example_prompts}. We do not claim that word perturbations are representative of all distribution shifts. In fact, we do not claim that word perturbations are representative of \textit{any} real-world distribution shift. Rather, we use word perturbations to allow analysis of OOD patterns in reward models where we can explicitly measure and induce structured OOD-ness. We further note that for our far-OOD experiments (where the perturbation percentage is high), it's unclear if the preference ranking label would still hold - and this is presented solely to to show the OOD patterns of reward models taken to the extreme.

All artificial perturbation experiments are run across 10 trials, where each trial corresponds to a different set of random perturbations. We plot the resulting experimental means as solid lines and standard deviations as error bars.

\subsubsection{Reward Model Performance Results Under Artificial Distribution Shift}

In Figure \ref{fig:dist_shift_acc_calibration}, we show performance of reward models in terms of accuracy and calibration under artificial distribution shift. We additionally present reward model confidence for context. We show that the accuracy of reward models in response to OOD prompts and responses degrades, similar to its classification counterparts. However, calibration is relatively unaffected by OOD prompts, similar to the findings in our experiments using natural distribution shifts; and calibration due to OOD responses follows a novel paradigm where accuracy drops more rapidly than confidence in near-OOD regions, then confidence drops rapidly and appropriately "catches up" with the drop in accuracy in far-OOD regions. As a result, calibration is excellent in response to far-OOD responses (interestingly, even better than ID response calibration), while calibration is poor in response to near-OOD responses.

We additionally note that these reward models are more susceptible to distribution shifts in responses than distribution shifts in prompts in terms of accuracy and confidence drops, as well as calibration changes. And we further note that even with completely random prompts, accuracy and confidence do not drop as much as expected, and calibration doesn't change as much as expected - suggesting reward models are relatively insensitive to OOD prompts.

Interestingly, the results from our lingual shift experiments in Table \ref{table:accuracy_degradation} more closely resemble the near-OOD results in Figure \ref{fig:dist_shift_acc_calibration} (specifically, around a $15\%$ perturbation probability) than the far-OOD results. Namely, non-english prompts cause slight ECE decreases and minimal accuracy drops when paired with english responses; and non-english responses when paired with either english or non-english prompts cause egregious ECE increases and accuracy drops. We conjecture this is due to cross-lingual correlations being learned in the pre-training stage of the reward model which utilizes a dataset which contains many languages. This perhaps allows the representations extracted from non-english inputs to be coerced into a multi-langual representation space before logit estimation. This would allow the fine-tuning leveraging the reward model to mostly focus on learning a mapping from multi-langual representations to reward scores, therefore leading non-english inputs to be interpreted as relatively similar to the english training set. Future work will further explore this phenomenon, especially as it relates to the possibility that such effect would show up in Large Language Models pre-trained on multi-lingual datasets and fine-tuned solely on one language.

\section{Detecting Distribution Shift In Prompts and Responses}

\subsection{A Simple Baseline To Detect OOD Prompts and Responses}
\label{sec:ood_detection}
To detect OOD prompts and responses, we can re-use the classification Energy Score but replace the classification logit function $\hat{L}^{class}$ with the reward score logit $\hat{L}^{rwd}$ as follows:
\begin{align*}
    S^{rwd}((p_i, (r_i^0, r_i^1)), \hat{L}^{rwd}) \\= -\log(e^{\hat{L}^{rwd}(p_i, r_i^0)} + e^{\hat{L}^{rwd}(p_i, r_i^1)})
\end{align*}

\subsection{Why Energy Score?}
 There exist many other Out-of-Distribution scores, but we use Energy Score here because it is shown to be an effective OOD score that takes inputs strictly from the outputs logits of the reward model. This is as opposed to other methods which use inference-time back-propagation \citep{liang2017enhancing, huang2021importance}, methods which modify to training \citep{hsu2020generalized}, sparsity methods which require access to intermediate representations \citep{sun2021react, djurisic2022extremely}, or methods which measure distance to intermediate representations \citep{sun2022out, lee2018simple} - all of which are inconvenient or impossible for reward models where you only have access to the output logits. We also use Energy Score over Maximum Softmax Probability (MSP) \citep{hendrycks2016baseline} - which is simply the model confidence $\hat{p}$ - because nearly all OOD benchmarks show that Energy Score outperforms MSP \citep{huang2021importance, sun2021react, sun2021effectiveness}. 

We do not introduce techniques which measure an explicit distance from an inference example to the reward model's training set, as reward model training datasets aren't always revealed or released.

\subsection{Performance of Our Baseline In Detecting OOD Prompts and Responses}

In Table \ref{table:natural_detection} and Figure \ref{fig:ood_detection}, we show the performance of our simple baseline presented in Section \ref{sec:ood_detection} in terms of its ability to detect natural and artificial OOD prompts and responses, respectively. As expected, OOD prompts and responses can be detected more easily as they become more OOD, as can be seen in Figure \ref{fig:ood_detection}. Moreover, it can detect shifts in responses more easily than shifts in prompts, as can be seen in both Table \ref{table:natural_detection} and Figure \ref{fig:ood_detection}. We further note that the lingual distribution detection results of Table \ref{table:natural_detection} closely resemble the near-OOD results of Figure \ref{fig:ood_detection} (also around a perturbation probability of $15\%$), similar to our earlier finding.

Results showing the inferiority of MSP in detecting artificial distribution shifts can be found in Appendix Figure \ref{fig:msp} - which is in agreement with the results of \citet{huang2021importance, sun2021react, sun2021effectiveness} in terms of finding MSP to be inferior to Energy Score in OOD detection.

\section{Conclusion}
In this work, we have provided a baseline study of reward models under distribution shift and introduced a method to detect OOD prompts and responses - with OOD responses detected more easily than OOD prompts in general. Specifically, we have shown that OOD prompts and responses induce accuracy drops in reward models - with OOD responses causing more egregious drops; and we have shown that the calibration of reward models is relatively unaffected by OOD prompts, while following a novel paradigm due to OOD responses where ID calibration is worse than far-OOD calibration but better than near-OOD calibration. Future work will explore if the same findings hold for different models and additional distribution shifts, such as style or subject.

\newpage
\bibliography{citations}

\newpage
\appendix
\onecolumn
\renewcommand{\thesection}{\Alph{section}}
\setcounter{section}{0}
\begin{center}
    \textbf{Appendix}
\end{center}

\begin{figure}[!htb]
    \centering
    \begin{subfigure}{0.4\textwidth}
        \centering
        \includegraphics[height=1.3in]{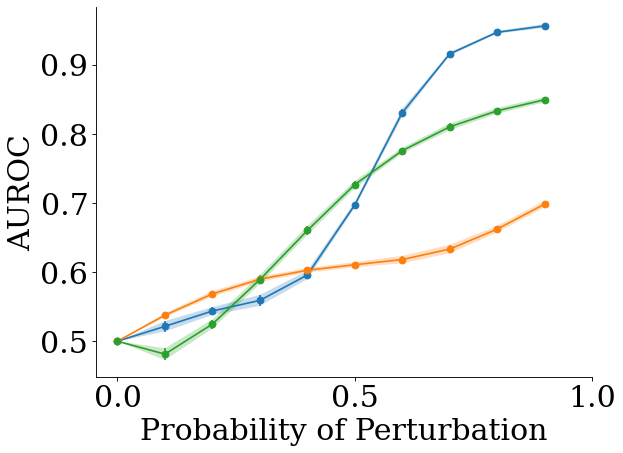}
    \end{subfigure}%
    ~ 
    \begin{subfigure}{0.6\textwidth}
        \centering
        \includegraphics[height=1.3in]{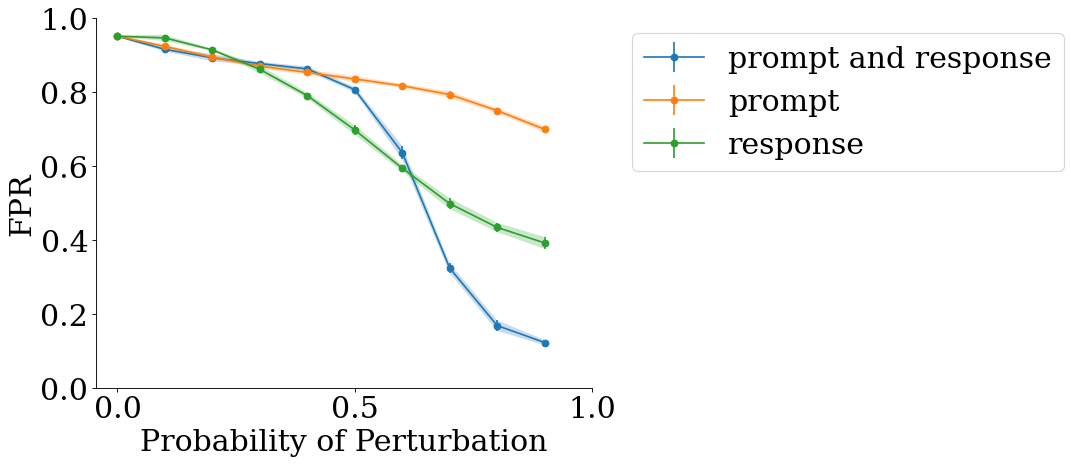}
    \end{subfigure}
    \caption{Performance of MSP in detecting artificial distribution shifts in prompts and responses, measured in terms of (\textbf{Left}) AUROC - where higher is better - and (\textbf{Right}) FPR@95 - where lower is better. Further right on the x-axis is further OOD.}
    \label{fig:msp}
\end{figure}

 \section{Example Prompts}
    \label{sec:example_prompts}
    \begin{enumerate}
        \item Here is an example prompt that was perturbed with 0\% or untouched: ``I am looking forward to hearing your thoughts about whether this relationship can be fixed or not."
        \item Here is the same example prompt but perturbed with a 25\% chance: ``I am looking forward to hearing your thoughts about whether this juryman vasty relationship can be trombidiid or not."
        \item The same example prompt but perturbed with a 50\% chance: ``am forward to divi-divi murem your thoughts about cochin this relationship can calcareous be fixed not."
        \item And finally the prompt perturbed with a 75\% chance: ``titian am looking forward hearing spherule thoughts waxflower about keratitis relationship booming be or alacrity cimetidine"
    \end{enumerate}

\section{Per-Language Results}
\label{sec:per_language_res}

Table \ref{table:per_language_acc} shows the per-language accuracy and ECE of the reward model in the lingual shift case, where the prompts and responses were translated.  The average of these results were shown in Table \ref{table:accuracy_degradation}. Table \ref{table:per_language_energy} shows the per-language AUROC and FPR@95 of the reward model, using the energy score as the OOD detection method. The average of these results were shown in Table \ref{table:natural_detection}.

\begin{table*}[!htb]
\centering
\begin{tabular}{llll}
\hline
Prompt Language      & Response Language         & Accuracy $\uparrow$         & ECE $\downarrow$\\ \hline
English &  English & 72.30\%   & 14.53\%  \\ 
\hline
English & French   & 65.52\%   &  20.37\%  \\
French & English   & 70.30\%   &  10.59\%   \\
French & French   &  64.46\%  &  19.87\%   \\
\hline
English & German   & 65.17\%   &  20.46\%  \\
German & English   &  70.18\%  &  10.72\%   \\
German & German   &  63.54\%  &  20.41\%    \\
\hline
English & Spanish   & 66.39\%   &  19.23\%  \\
Spanish & English   & 70.38\%   &  11.09\%   \\
Spanish & Spanish   &  65.32\%  &  19.10\%   \\

\hline
\end{tabular}
\centering
\caption{Comparison of reward model performances under lingual distribution shifts in prompt and response, with results shown per language. Here English is the in-distribution dataset, and all other languages are out-of-distribution. $\uparrow$ means higher is better and $\downarrow$ means lower is better.
}
\label{table:per_language_acc}
\end{table*}

\begin{table*}
\centering
\begin{tabular}{llll}
\hline
Prompt Language      & Response Language         & AUROC $\uparrow$         & FPR@95 $\downarrow$\\ \hline

English & French   & 74.26\%   &  70.03\%  \\
French & English   & 64.88\%   &  60.88\%   \\
French & French   &  79.27\%  &  57.04\%   \\
\hline
English & German   & 73.19\%   &  70.70\%  \\
German & English   &  63.12\%  &  72.17\%   \\
German & German   &  77.83\%  &  58.83\%    \\
\hline
English & Spanish   & 71.08\%   &  73.66\%  \\
Spanish & English   & 61.33\%   &  73.45\%   \\
Spanish & Spanish   &  74.25\%  &  64.78\%   \\

\hline
\end{tabular}
\centering
\caption{OOD detection results, using energy score, under lingual distribution shift, with results shown per language. Here English is the in-distribution dataset, and all other languages are out-of-distribution. $\uparrow$ means higher is better and $\downarrow$ means lower is better.
}
\label{table:per_language_energy}
\end{table*}

\end{document}